\documentclass{llncs}
\usepackage{amsmath}
\usepackage[a4paper]{geometry}
\usepackage{microtype}
\usepackage{siunitx}
\usepackage{booktabs}
\usepackage[colorlinks=false, pdfborder={0 0 0}]{hyperref}
\usepackage{cleveref}
\usepackage{latexsym}
\usepackage[english]{babel}
\usepackage[dvips]{graphicx}
\def\ni{\noindent}

\def\beq{\begin{equation}}
\def\eeq#1{\label{#1}\end{equation}}
\long\def\COMMENT#1\ENDCOMMENT{\message{(Commented text...)}\par}

\def\and{ \ \wedge}

\begin{document}
\title{Qsmodels: ASP Planning in Interactive Gaming Environment%
\thanks{Work supported by the Information
Society Technologies  programme of the European Commission, Future and
Emerging Technologies under the IST-2001-37004 WASP project.}
}

\author{Luca Padovani\inst{1} \and Alessandro Provetti\inst{2}}

\institute{
$M^2AG$: Milan-Messina Action Group\\
DSI--Universit\`a degli studi di Milano. Milan, I-20135 Italy\\
\email{luca@mag.dsi.unimi.it}.\\
\texttt{http://mag.dsi.unimi.it/}
\and
$M^2AG$: Milan-Messina Action Group\\
Dip. di Fisica--Universit\`a degli studi di Messina. Messina, I-98166 Italy\\
\email{ale@unime.it}\\
%\texttt{http://ale.unime.it/}
}

\maketitle

%%%%%%%%%%%%%%%%%%%%%%%%%%%%%%%%%%%%%%
\begin{abstract}
Qsmodels is a novel application of Answer Set Programming to interactive gaming environment. 
We describe a software architecture by which the behavior of a bot acting inside the Quake 3 Arena can be controlled by a planner. 
The planner is written as an Answer Set Program and is interpreted by the Smodels solver.
\end{abstract}

%%%%%%%%%%%%%%%%%%%%%%%%%%%%%%%%%%%%%%%%%%%%%%%%%%%%%%%%%%%%%

\ni
This article describes the Qsmodels project, which grew out of a graduation project \cite{Pad04} is currently under development.
The aim of this project is twofold. 

First, we want to demonstrate the viability of using \emph{Answer Set Programming} \cite{Bar03}  (ASP) in an interactive environment. 
The chosen environment is the \emph{Quake 3 Arena} (Q3A) game from \emph{id Software;}  recently most of the source codes have been released to the public. 
\emph{Q3A} is a \emph{first person shooter}: the player's goal is to kill enemies using weapons and upgrades found inside the game field (normally a labyrinth). 
The human-like enemies found within \emph{Q3A} are called \emph{BOTs.}
Like in the most computer games, Q3A bots behave according to the rules of a finite-state automaton (FSM) defined by expert game programmers.

The second objective is to implement and experiment with the high-level agent architecture described by Baral, Gelfond and Provetti \cite{BarGelPro97}. Such schema consists in the following loop: \emph{Observe--Select Goal--Plan--Execute.}

\par 
The Qsmodels architecture consists of two layers: a \emph{high level,} responsible for mid- and long-term planning, and a (\emph{low level}) in charge of plan execution and emergency state reactions. 
The high level has been developed mainly in ASP on the Smodels platform.
I.e., {\tt smodels} computes the \textit{answer sets} of a logic program which characterizes all successful plans of a given length, following the more or less standard encoding found in \cite{Bar03}. 
The computed answer set is passed to the low-level layer  that inspects it, extracts relevant syntactic informations and \textit{executes} the required actions.

\par 
The high-level layer of our project realizes a \emph{Q3A} agent which, starting with the knowledge about the game field similar to that of an intermediate-level human player, tries to beat his opponents by facing them only when in a better condition for the attack. 
To achieve this result, we have added to the planner a very simple learning system which keeps track of opponent's behavior in order to better guess future moves.

\par 
Even though we are still in the experimental phase, we submit that our architecture has several advantages over the traditional schema for the \textit{AI} part of games. 
Namely, our solution is easier to develop and keeps the AI at higher level of abstraction.
The easiness in development is reached by keeping the planning rules separated from the \textit{world model description} rules, so that they can be written even by AI beginners. 
Also, Qsmodels could be used for virtual-reality AI experiments; the use of a computer game as the laboratory environment allows choosing the level of abstraction of the physical model while, -at the same time- giving a useful visual feedback of agent's actions.

\par 
Finally, this project may help in evaluating the feasibility of using {\tt smodels} in near real-time applications and environments.
Indeed, we noticed a high computational demand to achieve realistic real-time behaviors.
More code analysis and optimization is in demand.

\vspace{-15pt}
\begin{multicols}{2}

\vspace{-30pt}
\begin{center}
{{\scriptsize $\ $}}\\
\includegraphics[scale=0.30]{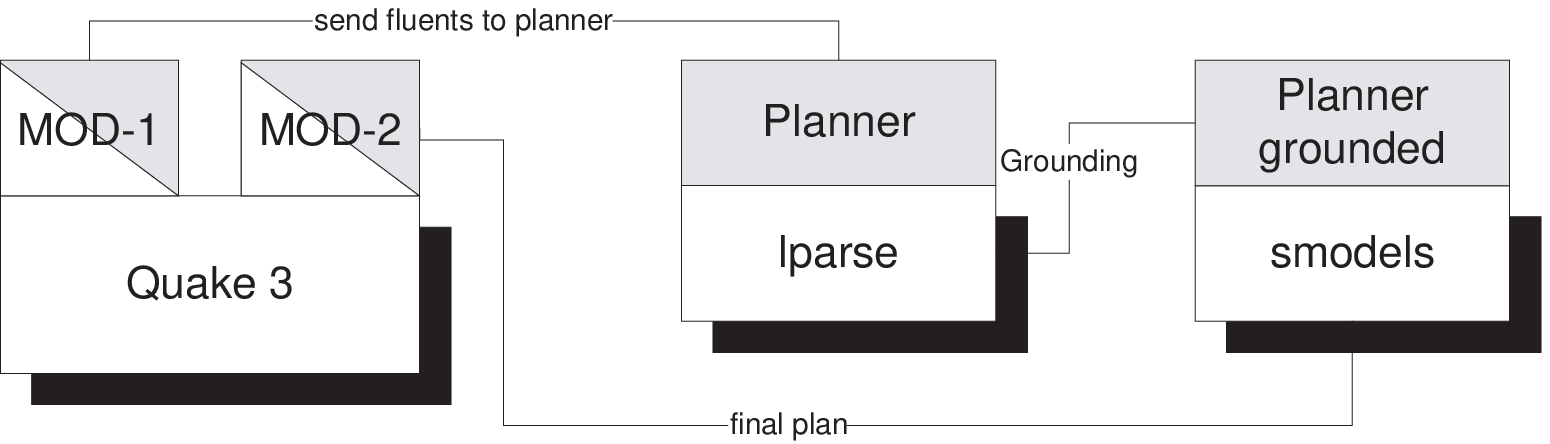}\\
{\scriptsize Figure 1. The Qsmodels Architecture\label{fig-schema-architettura}}
\end{center}

\vspace{-10pt}
\ni 
The Qsmodels software components can be divided in 3 parts: i) the ASP planner, ii) the \textit{Q3A} \textit{C++} interface, which implements \textit{sensing} and plan execution, and iii) the \textit{C++} low-level AI.

\end{multicols}

\vspace{-5pt}
\noindent
The execution model of Q3A, shown in Fig. \ref{fig-schema-architettura}, is summarized as follows. First, {\tt MOD-1} does the \textit{sensing} phase by inspecting some \textit{Q3A} memory areas; then, the computed informations is translated into high-level \textit{fluent values} and added to the planner. The planner is first grounded by {\tt lparse} then passed to {\tt smodels,} which computes one of its answer sets. 
Finally, {\tt MOD-2} extracts the plan from the answer set and executes it by calling the relative \textit{Q3A} \textit{traps}.

%%%%%%%%%%%%%%%%%%%%%%%%%%%%%%%%%%%%%%%%%%%%%%%%%%%%%%%%%%%%%
\subsection*{Methodology}
Our implementation required a lot of work and experiments in order to interface the existing software components, \textit{Q3A} and {\tt smodels}. 
The development of the agent required getting an in-depth knowledge of \textit{Q3A} internal functions, most of which are not documented. {\tt Smodels,} on the other hand, has been used 
as an external process, invoked by system calls. We are planning to switch to an API interface soon.

\par 
The two layers of Q3A work are executed concurrently: while the high level does the planning the low level is responsible for plan execution and reactive behavior in emergency situations. 
Events are deemed unforeseen when their occur makes the status of the domain incompatible with the assumptions made during the planning phase. 
It should be noticed that the low-level layer \textit{inherits} some powerful functionality from \textit{Q3A}, such as the \textit{combat} and \textit{shooting} actions, which are seen as atomic from the upper level.

\par 
To make our agent act realistically in its domain, we set the frequency of \textit{sensing} at ten times per second. 
This measure has to do with the way actions are executed: each action may consists of 
several repeated calls, until the \textit{goal} of the specific action is reached. 
So, since the plan execution is more associated with the \textit{frame frequency} of the game than with the plan actions, sensing needed to be executed even during action execution.

\par
To realize a \textit{reaction behavior,} we have introduced so-called \textit{pre-emption rules} which describe emergency behaviors in accordance with the environment and status of the plan. 
Pre-emption rules will be described in later sections.

%%%%%%%%%%%%%%%%%%%%%%%%%%%%%%%%%%%%%%%%%%%%%%%%%%%%%%%%%%%%%
\subsection*{Execution Cycle}
The  execution cycle of our application is shown in Figure 2. 
The first step is \textit{sensing,} where we access \emph{Q3A} memory searching for informations such as the agent state (position, health level \dots) and availability of bonuses (health and ammunition tokes).
Then, we check whether any emergency is happening, e.g., the agent is under attack, or he/she is facing the enemy, or he/she is behind the enemy etc.
If any of these situations holds, then we execute the \textit{pre-emption rules} to find 
those that apply to the present emergency and state. If no pre-emption rule applies then the execution cycle resumes.
 
\vspace{-10pt}
\begin{multicols}{2}
\vspace{-40pt}
	\begin{center}
{{\scriptsize $\ $}}\\
\includegraphics[scale=0.30]{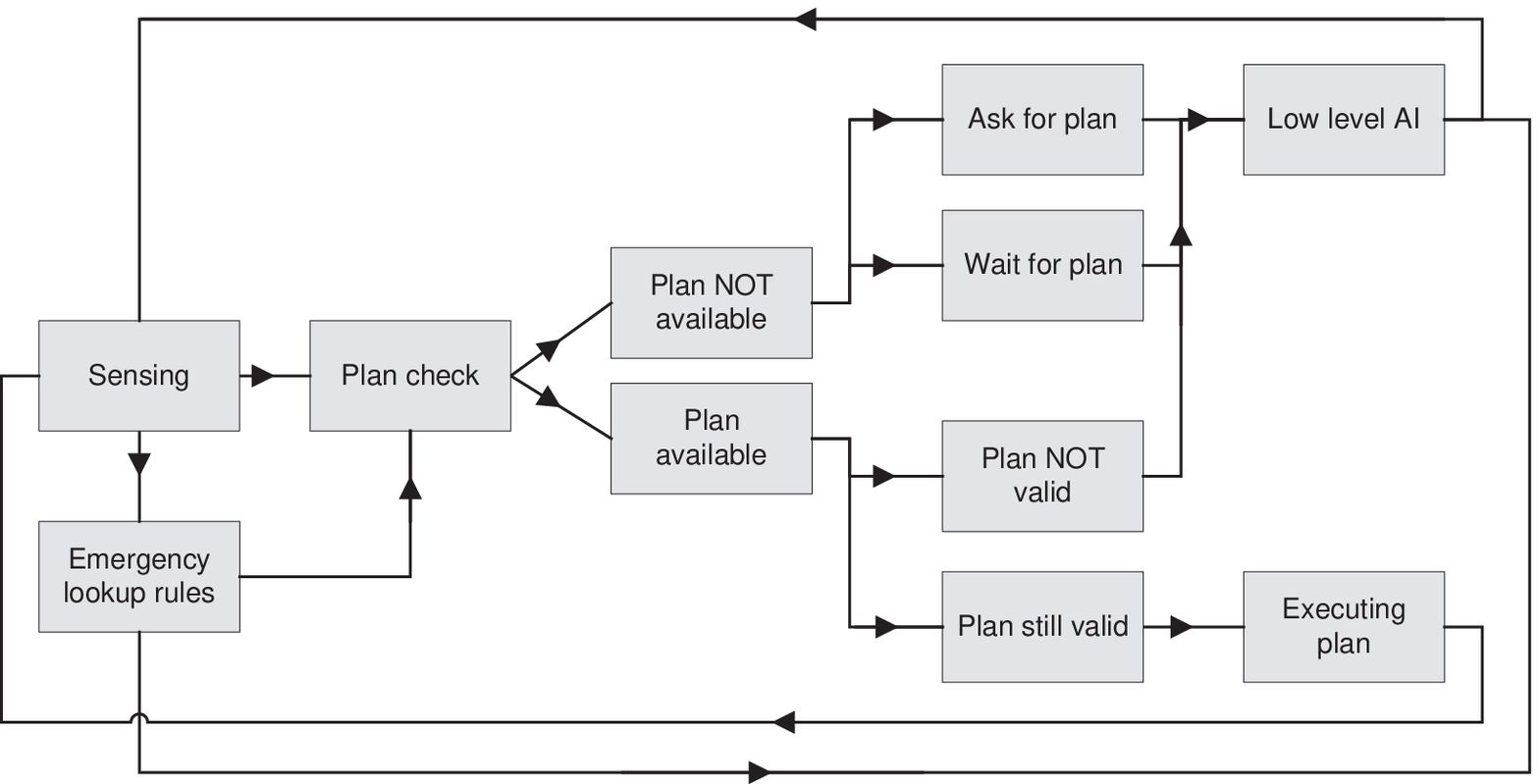}\\
{{\scriptsize Figure 2 - Execution Cycle.\label{fig-ciclo-esecuzione}}}
\end{center}

\ni
After sensing, if no emergency is detected we check whether a plan is currently available; 
if not, then we ask for a new one. Thus, we first translate the \textit{Q3A} memory states in \textit{fluents,} which are add as facts to the planner itself. Finally, we pass the augmented logic program to the external component
\end{multicols}

\vspace{-10pt}
\noindent 
\textit{QsmodelsServer,} which is in charge of the {\tt smodels} interface. 
The information embodied in the new fluents includes the agent's position, it's health and  weapons state and the positions of known active objects.
We include also a couple of atoms describing the \textit{last known enemy position} --- \textit{expected new enemy position} to try to find usual routes taken by the enemy.

\par 
If a plan is available, then we have to check if any of the agents or enemies actions have invalidated the assumptions made at the planning phase.
Indeed, since actions take some seconds to be executed and also the smodels computation can take several seconds, this situation would frequently happen, e.g., if the enemy takes a weapon that the agent was supposed to go get, the plan has to be invalidated since the weapon is not available anymore.

\par 
Let now consider plan execution. 
Each action available to the agent has been associated to a \textit{trap} to \textit{Q3A} system calls. 
The available actions are of course at very high level. 
This way, we have been able to reuse most of the basic AI work made by id Software: such as path finding and aiming.
So, the available actions are: {\tt move\_towards} -- {\tt pick\_health} -- {\tt pick\_ammo} -- {\tt attack} and {\tt elude}. 
All these actions except for {\tt attack} are variations of {\tt move\_towards}, since to get an object we have to reach it.  
Action {\tt attack} simply passes the control to the low-level AI in a  situation where the agent will certainly has to attack the enemy.

%%%%%%%%%%%%%%%%%%%%%%%%%%%%%%%%%%%%%%%%%%%%%%%%%%%%%%%%%%%%%
\subsection*{The use of pre-emption rules}
The game field of Q3A can be described as very dynamic. 
So, it would be unfeasible to recompute the plan each time some aspect of the environment changes. 
In this sense, the introduction of the so called \textit{Pre-emption rules} has probably been the most important step toward the realization of believable Q3A bots, 

\par 
Pre-emptive rules allows describing a high-level reaction system in which we specify the reaction behavior of the agent and let {\tt smodels} compute the actual reaction rules linked to the current plan. 
For each considered emergency situation and each time frame of the generated plan we get an appropriate reaction rule.

When an emergency happens our modification to \textit{Q3A} searches the corresponding rule (time and event) through the rules and executes the action inside the body of the rule. 
As a result, pre-emption rules dictate a behavior somewhat similar to that of a FSM.
However, in our case, the reaction is integrated in the planner and evaluated by the same 
inferential engine.
Therefore, we couple a time-consuming planning system for long-term reasoning to a more efficient reaction system for quick reactions.

%%%%%%%%%%%%%%%%%%%%%%%%%%%%%%%%%%%%%%%%%%%%%%%%%%%%%%%%%%%%%
\subsection*{Application Experience}
Our  testing platform consists of a set of \textit{Q3A} standard game levels in which our agent engages a duel against a human player. 
We require the game server (in our case an Intel P4 2.0GHz)to be run on a separate machine than that of the human player, due to the high computational power required by {\tt smodels}. 

The plan extraction phase can require up to 6/7 seconds, depending on the plan length. 
This delay is almost transparent to the human opponent, since during the planning our agent tries to hide. 
Should the agent meet the enemy then, the \textit{pre-emption rules} together with the low-level AI will make it act quickly, usually avoiding the confrontation.

\par 
In Qsmodels plans the last action is always {\tt attack} since the overall goal is to kill the enemy.
However, the last action seldom gets executed since when emergency situations happen the \textit{pre-emption rules} take control of the bot, canceling the residual part of the plan.

%%%%%%%%%%%%%%%%%%%%%%%%%%%%%%%%%%%%%%%%%%%%%%%%%%%%%%%%%%%%%
%\subsection*{Acknowledgments}

\vspace{-10pt}
%%%%%%%%%%%%%%%%%%%%%%%%%%%%%%%%%%%%%%%%%%%%%%%%%%%%%%%%%%%%%

%%%%%%%%%%%%%%%%%%%%%%%%%%%%%%%%%%%%%%
%%%%%%%%%%%%%%%%%%%%%%%%%%%%%%%%%%%%%%
\label{lastpage}

\begin{thebibliography}{}
%\bibitem[\protect\citename{ , }  19]{ }
% .,  (19)
\vspace{-7pt}
\bibitem{Bar03}%[Baral, 2003]
C.~Baral, 2003.
\newblock {\it Knowledge Representation, Reasoning and Declarative Problem Solving.}
Cambridge University Press.

\bibitem{BarGelPro97}%[Baral, 2003]
C.~Baral~M.~Gelfond~and~A.~Provetti, 1997.
\newblock {\it Representing Actions: Laws, Observations and Hypotheses.} Journal of Logic Programming, 31(1-3).

\bibitem{Pad04}%[Padovani, 2004]
L. Padovani, 2004.
\newblock \textit{Answer Set Programming in Interactive Gaming Environment.} 
\newblock Graduation project in Informatics (in Italian). University of Milan.
\newblock Available from \textit{http://mag.usr.dsi.unimi.it/}

\bibitem{systems}%[Systems]
Web location of the {\tt smodels} solver:\\
\newblock {\it http://www.tcs.hut.fi/Software/smodels/}

\end{thebibliography}
\end{document}